# Tensors Come of Age
## *Why the AI revolution will help HPC*

John L. Gustafson
Visiting Scientist at A*STAR and Professor at National University of Singapore
john.gustafson@nus.edu.sg

Lenore Mullin
Emeritus Professor at University at Albany, SUNY
lenore@albany.edu

**A Quick Retrospect**

Thirty years ago, parallel computing was coming of age. A bitter battle began between stalwart *vector computing* supporters and advocates of various approaches to *parallel computing*. IBM skeptic Alan Karp, reacting to announcements of nCUBE's 1024-microprocessor system and Thinking Machines' 65,536-element array, made a public $100 wager that **no one** could get a parallel speedup of over 200 on real HPC workloads. Gordon Bell softened that to an annual award for the **best** speedup, what we now know as the Gordon Bell Prize.

This year also marks the 30th Supercomputing Conference. At the first SC in 1988, Seymour Cray gave the keynote, and said he *might* consider combining up to 16 processors. Just weeks before that event, Sandia researchers had managed to get thousand-fold speedups on the 1024-processor nCUBE for several DOE workloads, but those results were awaiting publication.

The magazine *Supercomputing Review* was following the battle with interest, publishing a piece by a defender of the old way of doing things, Jack Worlton, titled "The Parallel Processing Bandwagon." It declared parallelism a nutty idea that would never be the right way to build a supercomputer. Amdahl's law and all that. A rebuttal by Gustafson titled "The Vector Gravy Train" was to appear in the next issue… but there was no next issue of *Supercomputing Review*. *SR* had made the bold step of turning into the first *online* magazine, back in 1987, with a new name.

Happy 30th Anniversary, *HPCWire*!

What better occasion than to write about another technology that is coming of age, one we will look back on as a watershed? That technology is *tensor computing*: Optimized multidimensional array processing using *novel arithmetic*. [1]

**Thank you, AI**

You can hardly throw a tchotchke on the trade show floor of SC'17 without hitting a vendor talking about artificial intelligence (AI), deep learning, and neural nets.

Google recently open-sourced its TensorFlow AI library and Tensor Processing Unit. Intel bought Nervana. Micron, AMD, ARM, Nvidia, and a raft of startups are suddenly pursuing an AI strategy. Two key ideas keep appearing:

- An architecture optimized for tensors
- Departure from 32-bit and 64-bit IEEE 754 floating-point arithmetic

*What's going on?* And is this relevant to HPC, or is it unrelated? Why are we seeing convergent evolution to the use of tensor processors, optimized tensor algebras in languages, and nontraditional arithmetic formats?

What's going on is that **computing is bandwidth-bound**, so we need to make *much* better use of the bits we slosh around a system. Tensor architectures place data closer to where it is needed. New arithmetic represents the needed numerical values using fewer bits. This AI-driven revolution will have a huge benefit for HPC workloads. Even if Moore's law stopped dead in its tracks, these approaches increase computing speed and cut space and energy consumption.

Tensor languages have actually been around for years. Remember APL and Fortran 90, all you old-timers? However, now we are within reach of techniques that can *automatically* optimize arbitrary tensor operations on tensor architectures, using an augmented compilation environment that minimizes clunky indexing and unnecessary scratch storage.[2] That's crucial for portability.

Portability suffers, temporarily, as we break free from standard numerical formats. You can turn float precision down to 16-bit, but then the shortcomings of IEEE format really become apparent, like wasting over 2,000 possible bit patterns on "Not a Number" instead of using them for numerical values. AI is providing the impetus to ask *what comes after floats*, which are awfully long in the tooth and have never followed algebraic laws. HPC people will someday be grateful that AI researchers helped fix this long-standing problem.

**The Most Over-Discovered Trick in HPC**

As early as the 1950s, according to the late numerical analyst Herb Keller, programmers discovered they could make linear algebra go faster by blocking the data to fit the architecture. Matrix-matrix operations in particular run best when the matrices are tiled into submatrices, and even sub-submatrices. That was the beginning of *dimension lifting*, an approach that seems to get re-discovered by every generation of HPC programmers. It's time for a "grand unification" of the technique.

**Level *N* BLAS**

The BLAS developers started in the 1970s with loops on lists (level 1), then realizing doubly nested loops were needed (level 2), then triply nested (level 3), and then LAPACK and SCALAPACK introduced blocking to better fit computer architectures. In other words, we've been computing with tensors for a long time, but not

admitting it! Kudos to Google for naming their TPU the way they did. What we need now is "level *N* BLAS."

Consider this abstract way of thinking about a dot product of four-element vectors:

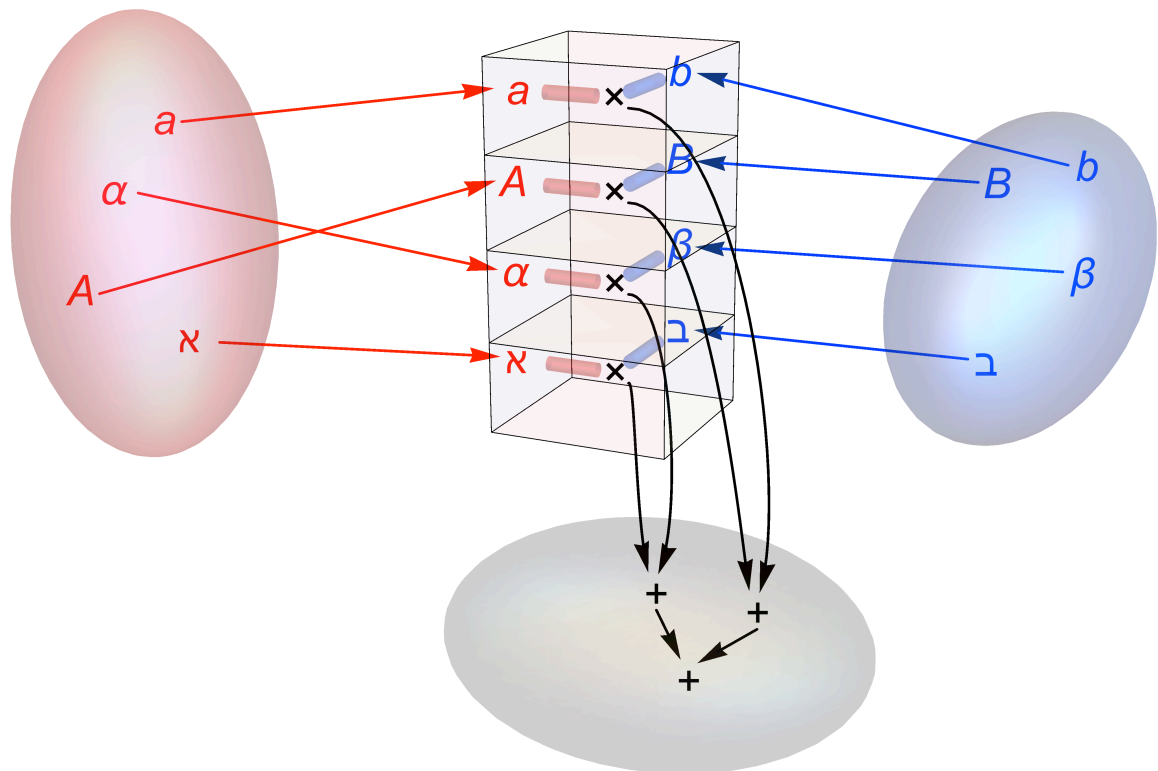

Notice the vector components are not numbered; think of them as a *set*, not a list, because that allows us to *rearrange* them to fit any memory architecture. The components are used once in this case, multiplied, and summed to some level (in this case, all the way down to a single number). Multiplications can be completely parallel if the hardware allows, and summation can be as parallel as binary sum reduction allows.

Now consider the same inputs, but used for 2-by-2 matrix-matrix multiplication:

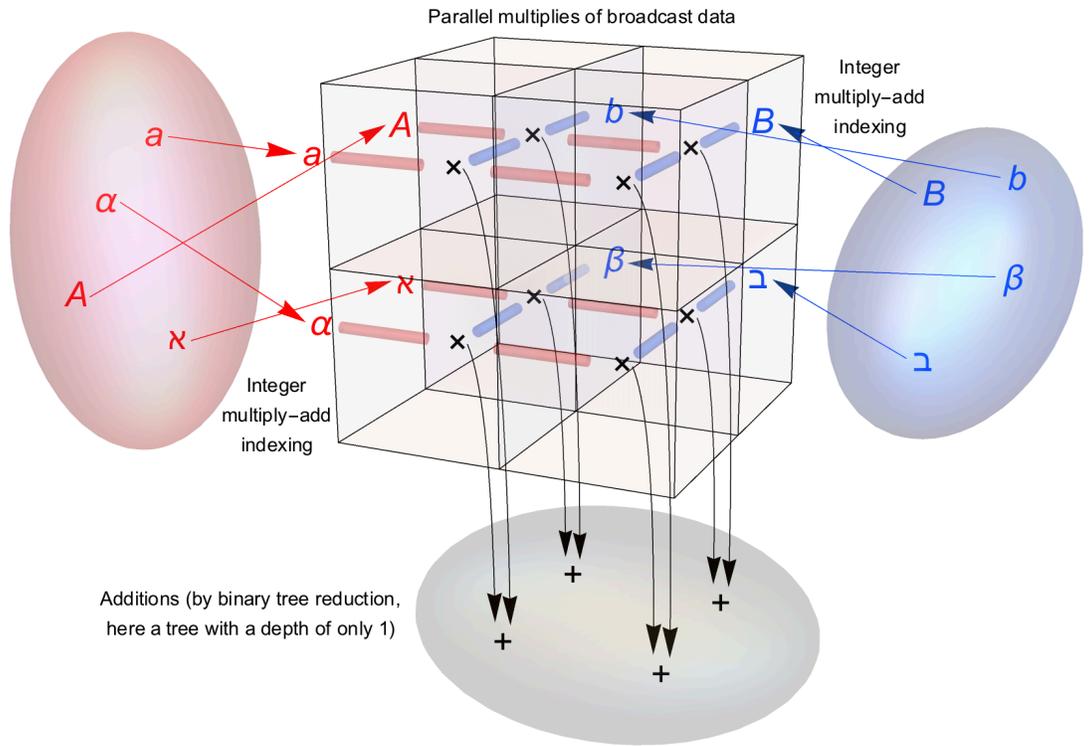

Each input is used twice, either by a broadcast method or re-use, depending on what the hardware supports. The summation is only one level deep this time.

Finally, use the sets for an *outer product*, where each input is used four times to create 16 parallel multiplications, which are not summed at all.

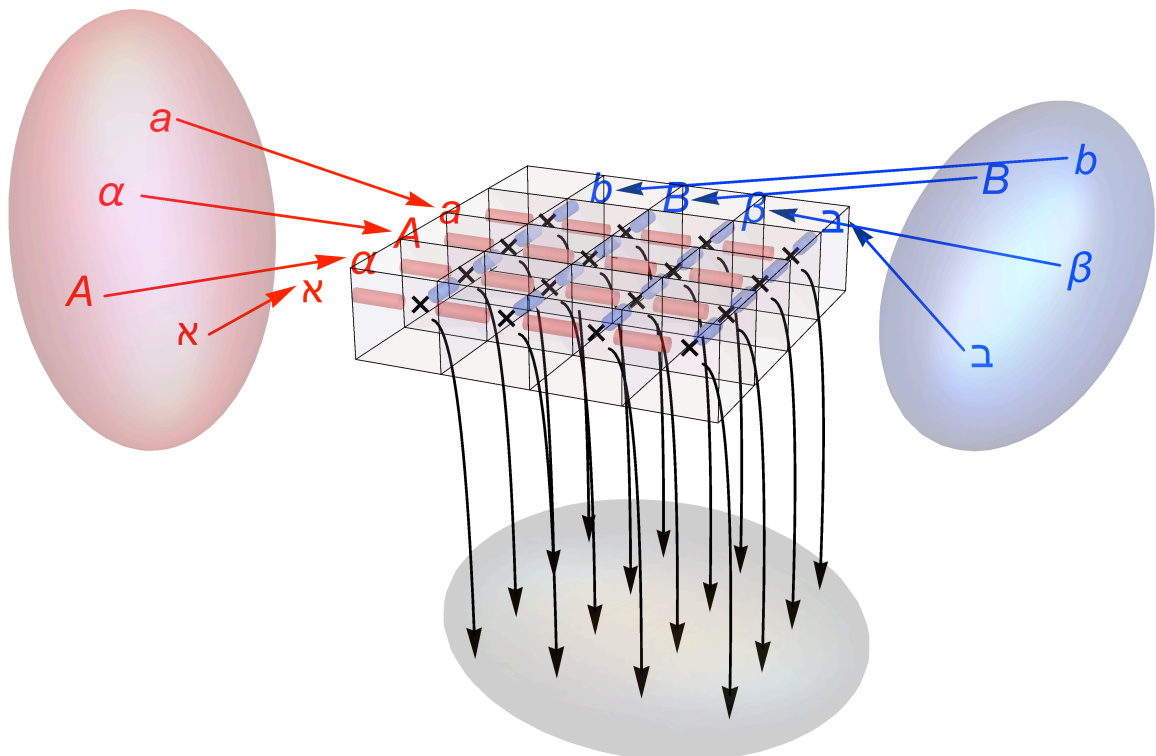

All these operations can be captured in a single unified framework, and that is what we mean by "Level *N* BLAS." The sets of numbers are best organized as *tensors that fit the target architecture* and its *cost functions*. A matrix really isn't two-dimensional in concept; that's just for human convenience, and semantics treat it that way. An algebra exists for index manipulation that can be part of the compiler smarts, freeing the programmer from having to worry about details like "Is this row-major or column-major order?"[3] Tensors free you from imposing linear ordering that isn't required by the algorithm and that impedes *optimal data placement*.

Besides linear algebra, tensors are what you need for Fast Fourier Transforms (FFTs), convolutions for signal and image processing, and yes, neural networks. Knowledge representation models like PARAFAC or CANDECOMP use tensors. Most people aren't taught tensors in college math, and tensors admittedly look pretty scary with all those subscripts. One of Einstein's best inventions was a shorthand notation that gets rid of a lot of the subscripts (because General Relativity requires tensor math), but it still takes a lot of practice to get a "feel" for how tensors work. The good news is, computer *users* don't have to learn that skill, and only a few computer *programmers* have to. There now exists a theory[4], and many prototypes[5], for handling tensors automatically. We just need a few programmers to make use of the existing theory of array indexing to build and maintain those tools for distribution to all.[6] Imagine being able to automatically generate a Fast Fourier Transform (FFT) without having to worry about the indexing! That's already been prototyped.[7]

Which leads us to another HPC trend that we need for architecture portability…

**The Rise of the Installer Program**

In the old days, code development meant edit, compile, link, and load. Nowadays, people never talk about "linkers" and "loaders." But we certainly talk about *precompilers*, *makefiles* and *installer programs*. We've also seen the rise of just-in-time compilation in languages like Java, with system-specific byte codes to get both portability and sometimes, surprisingly high performance. The nature of who-does-what has changed quite a bit over the last few decades. Now, for example, HPC software vendors cannot ship a binary for a cluster supercomputer because they cannot know which MPI library is in use; the installer links that in.

The compiler, or preprocessor, doesn't have to guess what the target architecture is; it can instead specify *what* needs to be done, but not *how*, stopping at an intermediate language level. The *installer* knows what the costs are of all the data motions in the example diagrams above, and can predict precisely what the cost of a particular memory layout is. What you can predict, you can optimize. The installer takes care of the *how*.

James Demmel has often described the terrible challenge of building a ScaLAPACK-like library that gets high performance for all possible situations. Call it "The

Demmel Dilemma." It appears we are about to resolve that dilemma. With tensor-friendly architectures, and proper division of labor between the human programmer and the preprocessor, compiler, and installer, we can look forward to a day when we **don't** need 50 pages of compiler flag documentation, or endless trial-and-error experimentation with ways to lay out arrays in storage that is hierarchical, parallel, and complicated. Automation is feasible, and essential.

**The Return of the Exact Dot Product**

There is one thing we've left out though, and it is one of the most exciting developments that will enable all this to work. You've probably never heard of it. It's the *exact dot product* approach invented by Ulrich Kulisch, back in the late 1960s, but made eminently practical by some folks at Berkeley just this year.[8]

With floats, because of rounding errors, you will typically get a different result when you change the way a sum is grouped. Floats disobey the associative law: $(a + b) + c$, rounded, is not the same as $a + (b + c)$. That's particularly hazardous when accumulating a lot of small quantities into a single sum, like when doing Monte Carlo methods, or a dot product. Just think of how often a scientific code needs to do the sum of products, even if it doesn't do linear algebra. Graphics codes are full of three-dimensional and two-dimensional dot products. Suppose you could calculate sums of products *exactly*, rounding only when converting back to the working real number format?

You might think that would take a huge, arbitrary precision library. It doesn't. Kulisch noticed that for floating-point numbers, a fixed-size register with a few hundred bits suffices as scratch space for perfect accuracy results even for vectors that are billions of floats long. You might think it would run too slowly, because of the usual speed-accuracy tradeoff. **Surprise:** It runs 3–6 times *faster* than a dot product with rounding after every multiply-add. Berkeley hardware engineers discovered this and published their result just this summer. In fact, the exact dot product is an excellent way to get over 90 percent of the peak multiply-add speed of a system, because the operations *pipeline*.

Unfortunately, the exact dot product idea has been repeatedly and firmly rejected by the IEEE 754 committee that defines how floats work. Fortunately, it is an absolute requirement in *posit arithmetic*[9] and can greatly reduce the need for double precision quantities in HPC programs. Imagine doing a structural analysis program with 32-bit variables throughout, yet getting 7 correct decimals of accuracy in the result, guaranteed. That's effectively like doubling bandwidth and storage compared to the 64-bits-everywhere approach typically used for structural analysis.

**A Scary-Looking Math Example**

If you don't like formulas, just skip this. Suppose you're using a conjugate gradient solver, and you want to evaluate its kernel as fast as possible:

$$\alpha_k := \frac{\mathbf{r}_k^T \mathbf{r}_k}{\mathbf{p}_k^T \mathbf{A} \mathbf{p}_k}$$

$$\mathbf{x}_{k+1} := \mathbf{x}_k + \alpha_k \mathbf{A} \mathbf{p}_k$$

A theory exists to mechanically transform these formulas to a "normal form" that looks like this:

$$\mathbf{X}[n+k] \equiv \mathbf{X}[k] + \mathbf{P}[k] \times \left( \sum_{j=0}^{n-1} \mathbf{R}[j]^2 \right) / \left( \sum_{i=0}^{n-1} \sum_{j=0}^{n-1} \mathbf{P}[i] \times \mathbf{A}[j + i \times n] \times \mathbf{P}[j] \right)$$

That, plus hardware-specific information, allows automatic data layout that minimizes indexing and temporary storage, and maximizes locality of access for any architecture. And with novel arithmetic like posits that supports the exact dot product, you get a *bitwise identical result* no matter how the task is organized to run in parallel, and at near-peak speed. Programmers won't have to wrestle with data placement, nor will they have to waste hours trying to figure out if the parallel answer is different because of a *bug* or because of *rounding errors*.

**What People Will Remember, 30 Years from Now**

By 2047, people may look back on the era of IEEE floating-point arithmetic the way we now regard the EBCDIC character set used on IBM mainframes (which many readers may never have heard of, but it predates ASCII). They'll wonder how people ever tolerated the lack of repeatability and portability and the rounding errors that were indistinguishable from programming bugs, and they may reminisce about how people wasted 15-decimal accuracy on every variable as insurance, when they only needed four decimals in the result. Not unlike the way some of us old-timers remember "vectorizing" code in 1987 to get it to run faster, or "unrolling" loops to help out the compiler.

Thirty years from now, the burden of code tuning and portability for arrays will be back where it belongs: on the computer itself. Programmers will have long forgotten how to tile matrices into submatrices because the compiler-installer combination will do that for tensors for any architecture, and will produce bitwise-identical results on all systems.

The big changes that are permitting this watershed are all happening *now*. This year. These are exciting times! □

---

[1] A. Acar et al., "Tensor Computing for Internet of Things," *Dagstuhl Reports*, Vol. 6, No. 4, 2016, Schloss Dagstuhl–Leibniz-Zentrum fuer Informatik,